\documentclass[letterpaper]{article} 
\usepackage{aaai25}  
\usepackage{times}  
\usepackage{helvet}  
\usepackage{courier}  
\usepackage[hyphens]{url}  
\usepackage{graphicx} 
\usepackage{natbib}  
\usepackage{caption} 
\usepackage{multicol}
\usepackage{multirow}
\usepackage{booktabs} 
\usepackage{amsmath}
\usepackage{xcolor}
\usepackage{makecell}
\usepackage{tcolorbox}

\usepackage{amsmath,amsfonts,bm}
\usepackage{graphicx}
\usepackage{sidecap}
\usepackage{mathrsfs}
\usepackage{multirow, multicol}
\usepackage{caption}

\usepackage{booktabs} 
\usepackage{enumitem}
\usepackage{longtable}

\def\eqref#1{equation~\ref{#1}}

\def\1{\bm{1}}

\DeclareMathAlphabet{\mathsfit}{\encodingdefault}{\sfdefault}{m}{sl}
\SetMathAlphabet{\mathsfit}{bold}{\encodingdefault}{\sfdefault}{bx}{n}

\def\rr{\textcolor{red}}
\def\bb{\textcolor{blue}}
\definecolor{ggg}{RGB}{26,179,0}

\usepackage[ruled, linesnumbered, lined]{algorithm2e}
\nocopyright 

\title{Self-Instructed Derived Prompt Generation Meets In-Context Learning: Unlocking New Potential of Black-Box LLMs}

\author {
    Zhuo Li\textsuperscript{\rm 1,\rm 2}\equalcontrib,
    Yuhao Du\textsuperscript{\rm 1,\rm 2}\equalcontrib,
    Jinpeng Hu\textsuperscript{\rm 3},
    Xiang Wan\textsuperscript{\rm 1},
    Anningzhe Gao\textsuperscript{\rm 1}\thanks{Corresponding author.}
}
\affiliations {
    \textsuperscript{\rm 1}Shenzhen Research Institute of Big Data\\
    \textsuperscript{\rm 2}The Chinese University of Hong Kong, Shenzhen\\
    \textsuperscript{\rm 3}Hefei University of Technology\\
    \{zhuoli3, yuhaodu\}@link.cuhk.edu.cn, jinpenghu@hfut.edu.cn, \{wanxiang, gaoanningzhe\}@sribd.cn
}

\usepackage{bibentry}

\begin{document}

\maketitle

\begin{abstract}

Large language models (LLMs) have shown success in generating high-quality responses. In order to achieve better alignment with LLMs with human preference, various works are proposed based on specific optimization process, which, however, is not suitable to Black-Box LLMs like GPT-4, due to inaccessible parameters. In Black-Box LLMs case, their performance is highly dependent on the quality of the provided prompts. Existing methods to enhance response quality often involve a prompt refinement model, yet these approaches potentially suffer from semantic inconsistencies between the refined and original prompts, and typically overlook the relationship between them. To address these challenges, we introduce a self-instructed in-context learning framework that empowers LLMs to deliver more effective responses by generating reliable derived prompts to construct informative contextual environments. Our approach incorporates a self-instructed reinforcement learning mechanism, enabling direct interaction with the response model during derived prompt generation for better alignment. We then formulate querying as an in-context learning task, using responses from LLMs combined with the derived prompts to establish a contextual demonstration for the original prompt. This strategy ensures alignment with the original query, reduces discrepancies from refined prompts, and maximizes the LLM’s in-context learning capability. Extensive experiments demonstrate that the proposed method not only generates more reliable derived prompts but also significantly enhances LLMs' ability to deliver more effective responses, including Black-Box models such as GPT-4.
\end{abstract}

%

\section{Introduction}
The emergence of Large Language Models (LLMs) has significantly advanced the field of Natural Language Processing (NLP), achieving remarkable results across various tasks~\cite{openai2023gpt,srivastava2022beyond,brown2020language,touvron2023llama,devlin2018bert}. The success of these models is highly dependent on the quality of the input prompts, as ambiguous or insecure prompts can lead to low-quality and unreliable responses~\cite{zhou2022large,zamfirescu2023johnny,liu2023pre}. 
Additionally, the high training costs associated with the large parameter sizes of these LLMs make it challenging to fine-tune for alignment when handling downstream tasks. In particular, for \textbf{Black-Box} LLMs, training is impossible. Therefore, utilizing better prompts to guide the models in generating the desired outputs has become an effective approach in applying LLMs to various tasks.

To this end, several approaches are proposed to find the optimal prompts that can generally promote LLMs across various tasks. A popular approach that improves the quality of the prompts is prompt engineering, which aims to refine the prompts with templates designed manually~\cite{tam2021improving,reynolds2021prompt}. However, this approach is limited in specific scenarios and requires the creation of new templates when transitioning to a new scenario, resulting in a high manual workload and limited usability~\cite{webson2021prompt}. Therefore, some methods are proposed to rewrite prompts, which usually leverage the powerful capabilities of LLMs (e.g., GPT-4) by directly asking the model to refine the prompts~\cite{zhou2022large, fernando2023promptbreeder}, or by training a dedicated prompt refinement model to enhance prompt quality~\cite{cheng2023black, deng2022rlprompt,kong2024prewrite}. 


\begin{table*}[h]
\centering
\resizebox{1\textwidth}{!}{
\begin{tabular}{c|c|l}
\toprule
\multirow{4}{*}{Case 1} & Original Prompt & Human knowledge is a collection of definitions and observations. What is your opinion about this proposition ? \\ 
&Refined Prompt& What is the foundation of human knowledge and how do we acquire and organize it? \\\cmidrule{2-3}
&Justification & \makecell[l]{\textcolor{blue}{The refined prompt diverges significantly from the original prompt by altering the focus and depth of inquiry.}} \\ \midrule \midrule 
\multirow{4}{*}{Case 2} &Original Prompt & Describe the \textcolor{red}{health benefits} of green tea. \\ 
&Refined Prompt& Discuss the \textcolor{red}{antioxidant properties} of green tea and its role in \textcolor{red}{preventing cancer}. \\ \cmidrule{2-3}
&Justification & \makecell[l]{\textcolor{blue}{The refined prompt loses context.} The original prompt asks about the health benefits of green tea in general, \\while the refined one focuses only on its antioxidant properties and role in cancer prevention, ignoring other benefits.} \\   \midrule \midrule 
\multirow{4}{*}{Case 3} &Original Prompt & Generate a product idea for a mobile application. \\ 
&Refined Prompt& Generate a product idea for a mobile application \textcolor{red}{that helps users meet dietary goals through personalized nutrition and meal planning.} \\ \cmidrule{2-3}
&Justification & \makecell[l]{\textcolor{blue}{The refined prompt overly narrows the focus compared to the original prompt.} It limits the potential product ideas to only those related \\to a specific topic, neglecting other innovative possibilities within the mobile application space.} 
\\  \bottomrule
\end{tabular} 
}
\caption{Failed refined prompt cases. Case 1 is from directly asking Llama3-Instruct~\cite{touvron2023llama} for a prompt refinement. Both Case 2 and Case 3 are from BPO~\cite{cheng2023black}. \textit{Justification} provides an analysis from GPT-4 on how the refined prompts fail to serve as an effective evolution compared with the original ones. More cases are provided in supplementary materials.}
\label{refined_problem}
\end{table*}

While these methods have achieved significant results, several challenges may limit their practical use on downstream tasks. For example, previous methods often require extensive data collection~\cite{cheng2023black, huang2024enhancing} and complex template design~\cite{webson2021prompt}, and the lack of interaction with the queried model (e.g., response model) that is used in downstream tasks during rewritten can result in prompts not fully compatible with it~\cite{deng2022rlprompt,cheng2023black}. Moreover, the prompt rewriting process may introduce semantic inconsistencies between the refined and original prompts, potentially leading to less effective responses, as shown in Tab.~\ref{refined_problem}.
Given these considerations, we raise the following question:

\textit{Is there a more effective way to prompt the response model (e.g., Black-Box GPT-4) than solely generating a refined prompt?}

Based on the success of in-context learning (ICL) with LLMs, which improves model performance with additional closely related demonstrations~\cite{dong2022survey}, we propose a novel framework to stimulate LLMs to generate more helpful and reliable responses by automatically constructing an informative in-context environment for the original prompt.
Instead of training a prompt refinement model that might alter the original intent, we design a \textit{\textbf{derived prompt}} generation model optimized by a self-instructed reinforcement learning (RL) objective. By integrating the response model into the RL training process, our method eliminates the need for training data collection and ensures closer alignment between the derived prompt and the response model.
Moving beyond directly querying LLMs with the derived prompt as the final response, we example the derived prompt-response pair as a semantically relevant and high-quality demonstration~\cite{liu2021makes} for the original prompt. This approach preserves the original prompt's information while leveraging the benefits of high-quality \textbf{\textit{in-context environment}}, which contains semantically similar information and response. Therefore, it effectively helps stimulate the LLMs' inherent knowledge, resulting in higher-quality and more helpful responses for original prompt.
Extensive experiments on various downstream datasets show significant performance improvements in response quality than prompt refinement methods, where our method also improves Black-Box models (e.g., GPT-4) in producing higher-quality response, demonstrating a promising and interpretable paradigm for aligning LLMs without the modifications. We highlight our advantages as follows:
\begin{itemize}
\item \textbf{Novel framework and preserving original intent.} We introduce a new framework to align LLMs with human preference which can be applied to \textbf{Black-Box} models. Our framework automatically constructs informative in-context learning (ICL) environments to stimulate LLMs for higher-quality and more helpful responses, which is achieved by a high-quality demonstration for the original prompt based on the derived prompt, preserving its original intent of user prompt while leveraging the benefits of in-context learning.
\item \textbf{Data collection free and Better alignment.} We propose to a self-instructed RL objective for effective derived prompt generation model optimization, where we integrate the response model into this training process, eliminating the need for extensive training data collection and better aligning derived prompts with it.
\item \textbf{Significant improvements.} Extensive experiments on various downstream datasets demonstrate significant improvements in response quality over existing prompt refinement methods, including enhancements in Black-Box models (e.g., GPT-4), showcasing a promising and interpretable paradigm for aligning LLMs without modifications.

\end{itemize}


\section{Background}
\paragraph{Supervised Fine-Tuning} Supervised Fine-Tuning (SFT) with annotated text descriptions is widely used to adapt LLMs into downstream tasks. Given prompt-response pairs ${(x,y)}$ sampled from a distribution $\mathcal{D}$, SFT objective function is defined as:
\begin{equation}
    \mathcal{L}_{\text{SFT}} = -\mathbb{E}_{(x,y) \sim \mathcal{D}} \left[ \sum_{i} \text{log}\  \pi_{\text{SFT}}(y_i|x,y_{<i})\right],\label{sft_obj}
\end{equation}
where $\pi$ indicates a LLM policy and $y_{<i}$ refers to all tokens before the $i-$th token in response $y$. In prompt rewritten task, $x$ and $y$ usually indicate original and refined prompt, respectively. For example,~\citet{cheng2023black} collects various original-refined prompt pairs as $(x,y)$ with the help of GPT-4 and then designs to optimize a prompt refinement model by minimizing Eq.~\ref{sft_obj}.
\paragraph{Reinforcement Learning from Human Feedback} Reinforcement Learning from Human Feedback (RLHF) is another effective tuning method for improving the alignment of LLMs with human preferences, which typically involves two steps: reward modeling and RL training. In reward modeling, a reward model $\mathcal{R}$ is designed to measure response quality to an input prompt: $\mathcal{L}_{\text{Reward}}=-\mathbb{E}_{(x,y_c,y_r) \sim \mathcal{D}} [ \text{log}(\sigma(\mathcal{R}(x,y_c) - \mathcal{R}(x,y_r))) ],$ where $y_c$ and $y_r$ indicate good and bad response, respectively. $\sigma$ is the sigmoid function. Generally, RL training uses the PPO algorithm~\cite{schulman2017proximal} with an additional Kullback–Leibler (KL) regularization as below:
\begin{equation}
    \mathcal{L}_{\text{RLHF}}=\mathbb{E}_{(x \sim \mathcal{D}, y \sim \pi_{\theta}(y|x))} \bigg [ \mathcal{R}(x,y) - \beta\ \text{log}\frac{\pi_{\theta}(y|x)}{\pi_{\text{SFT}}(y|x)} \bigg ],
\label{rlhf_obj}
\end{equation}
where $\beta > 0$ is a hyper-parameter that controls the influence of the KL penalty. 
Training a prompt rewritten model using Eq.~\ref{rlhf_obj} is still impractical because it requires collecting a dataset of original prompts and refinements, for obtaining a basic rewriter $\pi_{\text{SFT}}$ and a specific reward model which can evaluate the quality of generated refined prompt.

\begin{figure}[t]
    \centering
    \includegraphics[width=\linewidth]{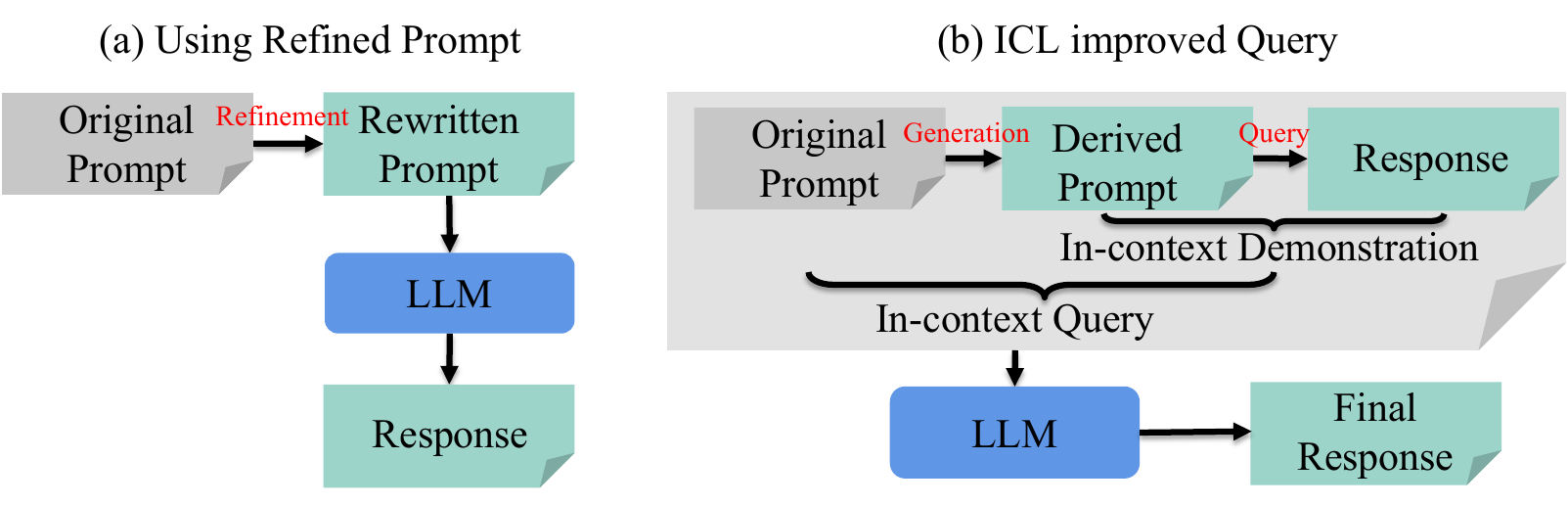}
    \caption{(a) Traditional methods directly replace the original prompt with a refined one, potentially risking semantic inconsistencies and ineffective responses. (b) Our method uses a derived prompt to create an in-context demonstration, ensuring high-quality responses while maintaining the integrity of the original prompt.}
    \label{fig:motivation}
\end{figure}

\section{Motivation}
As shown in Fig.~\ref{fig:motivation}(a), previous methods that directly replace the original prompt with a refined prompt to query LLMs often overlook valuable information contained in the original prompt, leading to ineffective responses due to essential semantic inconsistencies between the original and rewritten prompts. Fig.~\ref{fig:motivation}(b) gives a illustration to our methods, which addresses this issue by generating a derived prompt that is used to query the response model. The derived prompt and its response are then employed to construct a semantically similar, high-quality demonstration for the original prompt. This in-context query process promotes the model's in-context learning capabilities more effectively, ensuring high-quality responses while consistently querying the original prompt, thereby avoiding ineffective results.

\section{Methodology}

\subsection{Task Formulation}

We first define the concept of a derived prompt and its generation process. A derived prompt $x'$ is a transformation of the original prompt $x$ that maintains close relevance and shows improved expression, without necessarily being a rewritten or refined prompt. Our goal is to train an effective generation model $\pi$ that is initialized by a LLM (e.g., Llama3-Instruct) and can reliably produce a derived prompt $x' \sim \pi(\cdot|x)$ given original prompt. Consequently, a higher-quality response $y'$ can be generated by query a response model $\mathcal{M}$, still serving as a useful and effective feedback for $x$, due to their similar semantics. Generally, the model $\pi$ can be a frozen Black-Box model (e.g., GPT-4) or a learnable model (e.g., the Llama family~\cite{touvron2023llama}). In this research, we focus on optimizing a trainable model $\pi$ parameterized by $\theta \in \mathbb{R}^{d}$, with the goal of $x'$ to be more semantically derived to $x$ and better aligned with $\mathcal{M}$.

\subsection{Self-instructed RL for Derived Prompt Generation Model}

\begin{algorithm}[t]
\SetKwInOut{Input}{Input}
\SetKwInOut{Output}{Output}
\caption{Self-instructed RL for derived prompt generation model training.}
\Input{Training Dataset $\mathcal{D}$, DPG instruction $x_\text{DPG}$, a LLM initialized $\pi$, frozen response model $\mathcal{M}$ and reward model $\mathcal{R}$.}
\Output{Derived Prompt Generation Model $\pi_{\theta}$.}\label{alg:alg_1}

{Initialize $\pi_{\theta} \leftarrow \pi$}\;
{Initialize $\pi_{\text{ref}} \leftarrow \pi$}\;
\For{Random Sample a $x$ in $\mathcal{D}$}
{   
    {Obtain $X = \text{Concat}([x_\text{DPG}, x])$}\;
    {Prompt $\pi_{\theta}$ to obtain derived prompt $x' \sim \pi_{\theta}(X)$}\;
    {Prompt $\mathcal{M}$ to generate response $y' \sim \mathcal{M}(x')$}\;
    {Compute reward value by $\mathcal{R}(x', y')$}\;
    {Compute KL penalty by $\beta\ \text{log}\frac{\pi_{\theta}(x'|X)}{\pi_{\text{ref}}(x'|X)}$}\;
    {Update derived prompt generation model $\pi_{\theta}$ by maximizing $\mathcal{R}(x', y') - \beta\ \text{log}\frac{\pi_{\theta}(x'|X)}{\pi_{\text{ref}}(x'|X)}$}\;
}
\end{algorithm}


As mentioned in previous section, directly using Eq.~\ref{sft_obj} and~\ref{rlhf_obj} to optimize a derived prompt generation model $\pi_{\theta}$ typically introduces onerous data collection and lacks of alignment between derived prompts and response model, leading to suboptimal compatibility with downstream tasks. To address these issues, we propose a self-instructed RL objective for effective refinement model training.

Let $D=\{x_i\}_{i=1}^{N}$ denote a training data set of length $N$, which only includes the original prompts. Assume that we have a reliable reward model $\mathcal{R}$, which measures the response quality of the response model $\mathcal{M}$ when using the derived prompt $x'$. Instead of utilizing pre-collected $x'$ to optimize the model $\pi_{\theta}$ in an offline mode, we aim to leverage the reward for $\mathcal{M}$ response to $x'$ to direct the optimization, which can further generate desirable $x'$ that is better aligned with $\mathcal{M}$ (We freeze $\mathcal{M}$ during $\pi_{\theta}$ training since the parameters of $\mathcal{M}$ are inaccessible). To this end, we design to maximum the following objective:
\begin{equation}
\mathbb{E}_{(x \sim \mathcal{D}, x' \sim \pi_{\theta}(x'|x), y' \sim \mathcal{M}(x'))} \bigg [ \mathcal{R}(x', y') - \beta\ \text{log}\frac{\pi_{\theta}(x'|x)}{\pi_{\text{ref}}(x'|x)}\bigg],
\label{main_loss}
\end{equation}
where $x'$ is from the derived prompt generation model $\pi_{\theta}(x)$ and $y'$ is from the response model $\mathcal{M}(x')$, $\pi_{\text{ref}}$ is the reference model identically initialized by $\pi$ and $\beta$ is a hyper-parameter for stable model training. 

It is important to note that Eq.~\ref{main_loss} relies on the $\pi_{\theta}$ model having basic derived prompt generation capabilities. Given an original prompt $x$, $\pi_{\theta}$ can directly rewrite it instead of performing the task in $x$ (e.g., answering the question posed by $x$). Therefore, previous methods~\cite{kong2024prewrite, huang2024enhancing} necessitate an SFT stage to transform the pre-trained $\pi$ into a prompt rewriter $\pi_{\text{SFT}}$, which initializes both $\pi_{\theta}$ and $\pi_{\text{ref}}$, demanding extensive data collection of $(x, x')$ pairs.

Note that a well pre-trained $\pi_{\theta}$ model (e.g., Llama3-Instruct) should inherently possess basic capabilities of instruction-following~\cite{ouyang2022training} and paraphrasing. Therefore, we propose to use a derived prompt generation (DPG) instruction $x_\text{DPG}$ to overcome the necessity of SFT. Specifically, we manually design a $x_\text{DPG}$, which is provided in supplementary materials. As expected, $\pi_{\theta}$ can still generate a useful derived prompt $x'$ based its instruction-following capability:
\begin{equation}
\begin{aligned}
    X &= \text{Concat}([x_\text{DPG}, x]), \\
    x'&\sim \pi_{\theta}(X).
\end{aligned}
\label{obtain_derived_x}
\end{equation}

With the help of $x_\text{DPG}$, our method can effectively eliminate the need for data collection and additional training costs introduced by SFT. Additionally, by strategically leveraging the $\mathcal{M}$ into the training process of $\pi_{\theta}$, the generated derived prompts will be more in line with the preferences of the response model. Our final training objective is shown below:
\begin{equation}
\mathbb{E}_{(x \sim \mathcal{D}, x' \sim \pi_{\theta}(x'|X), y' \sim \mathcal{M}(x'))} \bigg [ \mathcal{R}(x', y') - \beta\ \text{log}\frac{\pi_{\theta}(x'|X)}{\pi_{\text{ref}}(x'|X)}\bigg].
\label{main_loss_1}
\end{equation}
We summarize the training process in Alg.~\ref{alg:alg_1}.


\subsection{Intent-consistency Oriented In-context Query Framework for Inference}
Although the derived prompt generation model trained using Eq.~\ref{main_loss_1} can produce higher-quality, semantically rich, and more compatible $x'$ for the response model, there remains a risk of semantic inconsistency and intent shift due to the uncontrollability of the generation process. Consequently, directly replacing the original prompt with the derived prompt would not be the optimal solution to make full use of it.

To address the mentioned issue and better activate the LLM's inherent knowledge, we propose a general in-context query framework to mitigate potential semantic inconsistencies, where we leverage high-quality, relevant in-context demonstrations derived from the original prompt. Therefore, this approaches can effectively enhance LLM ability to better respond to the user's original query. As shown in Alg.~\ref{alg:alg_2}, we construct an intent-consistent in-context query by filling the following template using $(x, x', y')$, where $x$, $x'$, $y'$ indicates the original prompt, its corresponding derived prompt and the LLM response to the derived prompt, respectively. This in-context query requires the better LLM respond to the original question $x$ by emulating the quality, style, and level of detail of the response $y'$ given to $x'$:


\begin{tcolorbox}
\begin{flushleft}
\textit{\#\#\# Question: \{Derived Prompt $x'$\}. } \\
\textit{\#\#\# Response: \{Response $y'$\}. } \\
\hspace*{\fill}\\
\textit{Given the above {Question} and {Response} as an example, emulate the way it responds to the question to reply to the following question:} \\
\hspace*{\fill} \\
\textit{\#\#\# Question: \{Original Prompt $x$\}.}
\end{flushleft}
\end{tcolorbox}

This in-context query framework ensures that the final response aligns more closely with the original intent included in $x$, while maintaining the enhanced response characteristics corresponding to the derived prompt. Therefore, it mitigates potential discrepancies between the refined prompt and the original prompt, leading to responses that are more helpful, reliable, and consistent with user expectations.

\begin{algorithm}[t]
\SetKwInOut{Input}{Input}
\SetKwInOut{Output}{Output}
\caption{Intent-consistency oriented in-context query framework for inference.}
\Input{Inference prompt $x$, DPG instruction $x_\text{DPG}$, derived prompt generation model $\pi$ and a LLM to be quried.}
\Output{Final response to user.}\label{alg:alg_2}
{Obtain $X = \text{Concat}([x_\text{DPG}, x])$}\;
{Generate refined prompt $x' \sim \pi(x'|X)$}\;
{Prompt LLM to obtain response $y' \sim \text{LLM}(x')$}\;
{Fulfill the in-context query template with $(x, x', y')$}\;
{Query $\text{LLM}$ and obtain final response}\;
\end{algorithm}

\subsection{Implementation Details}

\paragraph{Reward Model $\mathcal{R}$.} Recall that we use the generated response $y'$ corresponding to the derived prompt $x'$ to guide the optimization of $\pi_{\theta}$. To effectively evaluate the quality of generated pairs $(x', y')$, which determines the effectiveness of $\pi_{\theta}$, we employ a high-performance reward model that assigns a numerical score $\mathcal{R}(x, y)$ to the response $y$ of a model given a user query $x$, and reflects various aspects of response quality, including helpfulness, relevance, and coherence. This reward model is trained on hh-rlhf helpful dataset~\cite{bai2022training} based on GPT2-Large. While theoretically, any reasonable reward model $\mathcal{R}$ can also be used.

\paragraph{Response Model $\mathcal{M}$.} In the training process of the derived prompt generation model, we require a response model $\mathcal{M}$ that generates a response $y'$ to the derived prompt $x'$ for maximizing Eq.~\ref{main_loss_1}. Theoretically, $\mathcal{M}$ can be any LLMs capable of generating responses to given input prompts (e.g., the Llama family and GPT-4). In our experiment, we choose both White-Box model like Llama series and Black-Box model like GPT-4 as our query model.


\begin{table*}[!t]
\centering

\resizebox{1\textwidth}{!}{
\begin{tabular}{c|c|cc|ccc|ccc|ccc|ccc}
\toprule
\multirow{2}{*}{\LARGE$\pi_\theta$} & \multirow{2}{*}{Query Model} & \multicolumn{2}{c|}{Method} & \multicolumn{3}{c|}{Vicuna Eval} & \multicolumn{3}{c|}{BPO-test Eval} & \multicolumn{3}{c|}{Dolley Eval} & \multicolumn{3}{c}{Self-Instruct Eval} \\ 
 &  & A & B & A Win & B Win & Tie & A Win & B Win & Tie & A Win & B Win & Tie & A Win & B Win & Tie \\ \midrule \midrule
\multirow{11}{*}{Llama3} & \multirow{2}{*}{GPT-4} & OURS & OP & \textbf{90.0} & 3.8 & 6.2 & \textbf{71.0} & {24.5} & {4.5} & \textbf{80.5} & {15.5} & {4.0} & \textbf{76.2} & 5.6 & 18.3  \\
 &  & OURS & BPO & \textbf{88.8} & 7.5 & 3.7 & \textbf{74.0} & {25.5} & {7.5} & \textbf{71.0} & {27.0} & {2.0} & \textbf{71.4} & {21.0} & {7.6}  \\ \cmidrule{2-16}
 
 & \multirow{2}{*}{GPT-3.5} & OURS & OP & \textbf{93.8} & 2.5 & 3.7 & \textbf{77.5} & 19.5 & 3.0 & \textbf{79.5} & {6.0} & {4.5} & \textbf{84.5} & {9.9} & {5.6}  \\
 &  & OURS & BPO & \textbf{85.0} & 11.3 & 3.7 & \textbf{71.0} & {14.5} & {4.5} & \textbf{77.0} & {20.0} & 3.0 & \textbf{86.1} & 9.9 & 4.0   \\ \cmidrule{2-16}
 & \multirow{2}{*}{Llama3-8B} & OURS & OP & \textbf{82.5} & 15.0 & 2.5 & \textbf{65.5} & 30.0 & 4.5 & \textbf{59.0} & 34.0 & 7.0 & \textbf{78.9} & 15.1 & 6.0   \\
 &  & OURS & BPO & \textbf{81.3} & 35.0 & 1.2 & \textbf{63.0} & 34.0 & 3.0 & \textbf{51.0} & 47.0 & 2.0 & \textbf{75.8} & 22.2 & 2.0 \\ \cmidrule{2-16}
 & \multirow{2}{*}{Llama2-7B} & OURS & OP & \textbf{82.5} & 15.0 & 2.5 & \textbf{76.0} & 21.5 & 2.5 & \textbf{69.5} & 28.5 & 2.0 & \textbf{78.1} & 17.5 & 4.4 \\
 &  & OURS & BPO & \textbf{81.3} & 17.5 & 1.2 & \textbf{68.0} & 28.5 & 2.5 & \textbf{67.5} & 28.5 & 4.0 & \textbf{73.4} & 24.6 & 2.0 \\ \cmidrule{2-16}
 & \multirow{2}{*}{Qwen2-7B} & OURS & OP & \textbf{91.3} & 5.0 & 3.7 & \textbf{85.0} & 10.5 & 4.5 & \textbf{82.5} & 13.0 & 4.5 & \textbf{90.8} & 9.2 & 0 \\
 &  & OURS & BPO & \textbf{92.5} & 7.5 & 0.0 & \textbf{81.5} & 13.5 & 5.0 & \textbf{82.5} & 16.0 & 1.5 & \textbf{81.0} & 17.8 & 1.2 \\ \midrule\midrule
 \multirow{7}{*}{Llama2} 
 & \multirow{2}{*}{Llama3-8B} & OURS & OP & \textbf{85.0} & 12.5 & 2.5 & \textbf{67.5} & 28.0 & 4.5 & \textbf{61.5} & 35.0 & 3.5 & \textbf{77.0} & {13.9} & 9.1 \\
 & & OURS & BPO & \textbf{78.8} & 21.2 & 0.0 & \textbf{66.0} & 30.0 & 4.0 & \textbf{53.0} & 40.5 & 1.5 & \textbf{73.8} & 21.4 & 4.8 \\ \cmidrule{2-16}
 & \multirow{2}{*}{Llama2-7B} & OURS & OP & \textbf{86.3} & 12.5 & 1.2 & \textbf{74.5} & 22.0 & 3.5 & \textbf{74.5} & 23.0 & 2.5 & \textbf{84.1} & 13.0 & 2.9 \\
 &  & OURS & BPO & \textbf{85.0} & 10.0 & 5.0 & \textbf{66.0} & 29.5 & 4.5 & \textbf{67.5} & 32.0 & 0.5 & \textbf{72.2} & 25.9 & 9.9 \\ \cmidrule{2-16}
 & \multirow{2}{*}{Qwen2-7B} & OURS & OP & \textbf{91.3} & 8.7 & 0 & \textbf{83.0} & 13.0 & 4.0 & \textbf{85.0} & 11.0 & 4.0 & \textbf{92.1} & 7.5 & 0.4  \\
 &  & OURS & BPO & \textbf{85.0} & 15.0 & 3.0 & \textbf{79.0} & 17.5 & 3.5 & \textbf{78.5} & 20.5 & 1.0 & \textbf{84.5} & 14.7 & 0.8 \\ \midrule\midrule
 \multirow{7}{*}{Qwen2} 
 & \multirow{2}{*}{Llama3-8B} & OURS & OP & \textbf{81.3} & 17.5 & 1.2 & \textbf{74.5} & 23.0 & 2.5 & \textbf{59.5} & 36.5 & 4.0 & \textbf{77.8} & 12.7 & 9.5 \\
 &  & OURS & BPO & \textbf{78.8} & 21.2 & 0.0 & \textbf{69.5} & 26.0 & 4.5 & \textbf{54.5} & 42.5 & 3.0 & \textbf{69.8} & {27.8} & {2.4}  \\ \cmidrule{2-16}
 & \multirow{2}{*}{Llama2-7B} & OURS & OP & \textbf{92.5} & 7.5 & 0.0 & \textbf{74.0} & 23.5 & 2.5 & \textbf{71.0} & 23.0 & 6.0 & \textbf{84.1} & 11.1 & 4.8  \\
 &  & OURS & BPO & \textbf{78.8} & 21.2 & 0.0 & \textbf{67.0} & 29.5 & 3.5 & \textbf{73.0} & 23.5 & 3.5 & \textbf{77.4} & 21.0 & 1.6  \\ \cmidrule{2-16}
 & \multirow{2}{*}{Qwen2-7B} & OURS & OP & \textbf{96.3} & 2.5 & 1.2 & \textbf{86.5} & 8.0 & 5.5 & \textbf{82.0} & 13.0 & 5.0 & \textbf{84.5} & 14.7 & 0.8  \\
 &  & OURS & BPO & \textbf{95.0} & 3.8 & 1.2 & \textbf{67.0} & 15.0 & 4.0 & \textbf{78.0} & 19.0 & 3.0 & \textbf{91.7} & 8.0 & 0.3  \\ \midrule
\end{tabular}
}
\caption{A comprehensive comparison of OURS method, BPO and OP (Orginal Prompt) for different $\pi_{\theta}$ across four evaluation datasets. Query Model indicates the LLM used to response to the input prompt.}\label{main_res}
\end{table*}

\section{Experiments}
In this section, we conduct extensive experiments on various downstream datasets to comprehensively evaluate our method when compared with baseline methods. Ablation studies demonstrate the superior and necessity of our proposed self-instructed RL method for derived prompt generation, and also shows that ICL can serve as a flexible play-and-plug module to generally boosting existing methods to achieve better performance. We provide detailed training settings in supplementary materials.

\subsection{Experimental Setup}
\paragraph{Datasets.}To train a derived prompt generation model by Eq.~\ref{main_loss_1}, we utilize the BPO training dataset by following previous work~\cite{huang2024enhancing, cheng2023black}, which is constructed from four meticulously selected datasets and comprises 14K diverse samples. In order to comprehensively evaluate the performance of our method, we adopt wide-used instruction datasets for assessment, including Dolly Eval~\cite{conover2023free}, Vicuna Eval~\cite{chiang2023vicuna}, Self-Instruct Eval~\cite{wang2022self} and BPO test Eval~\cite{cheng2023black}. We provide detailed description in supplementary materials.

\paragraph{Derived Prompt Generation Model.}We particularly focus training an effective $\pi_\theta$ based on the popular LLMs such as LLaMA~\cite{touvron2023llama} and Qwen~\cite{yang2024qwen2}. Specifically, considering that our method relies on an instruction-following capability for derived prompt generation, we apply our method to to Llama2-chat~\cite{touvron2023llama}, Llama3-8B-Instruct~\cite{touvron2023llama} and Qwen2-7B-Instruct~\cite{yang2024qwen2}, which are all abbreviated as Llama3, Llama2 and Qwen2, respectively.

\paragraph{Queried Model.} A queried model is an frozen model used to generate responses to various prompt inputs, where we employ Llama2-chat (Llama2-7B), Llama3-Instruct (Llama3-8B), Qwen2-7B-Instruct (Qwen2-7B), GPT-3.5-turbo (GPT-3.5) and GPT-4o (GPT-4).

\paragraph{Baselines and Evaluation Metrics.}To demonstrate the effectiveness of our approach in help LLMs generate more effective response, we conduct a comprehensive comparison of the quality of the responses using original prompts and those refined through BPO method~\cite{cheng2023black}, due to all we mainly focus on better prompt Black-Box model to generate more helpful response. Following the common practice of employing LLMs for evaluation~\cite{zheng2024judging,wang2023pandalm}, we utilize GPT-4o to assess the impact of our method, where we adopt prompts from the MT-bench dataset~\cite{zheng2024judging}. To further ensure the fairness of our evaluation and to mitigate any potential position bias, we implement a random shuffling during each evaluation. In our results analysis, we mainly report Win Rates, which is computed by $(\text{A Win} - \text{B Win})\%$.


\subsection{Results}
We use OP to denote the responses obtained by the LLM from \textbf{O}riginal \textbf{P}rompt, where BPO and OD indicates that of \textbf{BPO} refined prompt and \textbf{O}ur \textbf{D}erived prompt, respectively. ICL represents the in-context learning environment. OURS indicates OD + ICL. As shown in Tab.~\ref{main_res}, our method shows an overall performance improvement than baseline methods, specifically for Black-Box model. Our method show multiple significant advantages:

\textbf{Stability Across Datasets:}
Our method demonstrates excellent performance and higher win rates across all four evaluation datasets compared to OP and BPO. The ICL queries generated by our method effectively promote the LLM to produce high-quality, more helpful, and comprehensive responses to the original questions, showcasing its robustness in handling various tasks, whether complex or simple.

\textbf{Consistency Across Models:}
Our method shows consistent performance improvement between different models. For example, in the Vicuna Eval dataset, the average OURS win rates in Llama3, Llama2 and Qwen2 are $73.6\%$, $76.3\%$, $80.8\%$ compared to OP, and those are $65.0\%$, $69.2\%$ and $68.8\%$ compared with BPO, respectively. These consistent higher results demonstrate that our method maintains high performance and robustness across a variety of underlying models, ensuring reliable results independent to models.


\textbf{Cross-Model Transferability:}
Tab.~\ref{main_res} also shows that the $\pi_\theta$ trained on Llama3 and constructed ICL queries can achieve better response quality on other models (e.g., Llama2 and Qwen2). For example, the model trained on Llama2 based on Dolley Eval has a win rate of $26.5\%$ on Llama3 and $74.0\%$ on Qwen2 compared with OD, and that are $12.5\%$ and $58.0\%$ compared with BPO. This suggests that our method has good cross-model transferability and can maintain high performance across different models.

\textbf{Improved Black-Box Model Performance:}
Our method significantly improved the response quality of black box models such as GPT-4 and GPT-3.5. For instance, when training on Llama3, the average win rates of our method on GPT-4 compared with OP and BPO are $67.1\%$ and $56.1\%$, respectively. 
And thoese on GPT-3.5 are $74.3\%$ and $69.9\%$, respectively.
The significant improvement indicates that our method effectively promote Black-Box LLM generate high-quality response. Similarly, in other datasets, the win rates of our method on GPT series models are also significantly higher than those of other methods.


\begin{table}[t]
\centering
\resizebox{0.47\textwidth}{!}{
\begin{tabular}{c|cc|ccc|ccc}
\toprule
\multirow{2}{*}{} & \multicolumn{2}{c}{Method} & \multicolumn{3}{c}{Vicuna Eval} & \multicolumn{3}{c}{Self-Instruct} \\
 & A & B & A Win & B Win & Tie & A Win & B Win & Tie \\ \midrule \midrule
\#1 & OD & OP & \textbf{78.8} & 11.2 & 10.0 & \textbf{66.3} & 15.5 & 18.2 \\
\#2 & BPO & OP & \textbf{68.8} & 15.0 & 16.2 & \textbf{66.3} & 21.4 & 12.3 \\ 
\#3 & OD & BPO & \textbf{72.5} & 21.3 & 6.2 & \textbf{42.1} & 8.7 & 49.2 \\ \cmidrule{2-9}
\#4 & BPO & OP & \textbf{68.8} & 15.0 & 16.2 & \textbf{66.3} & 21.4 & 12.3 \\
\#5 & BPO + ICL & OP & \textbf{76.3} & 11.3 & 12.4 & \textbf{69.4} & 21.4 & 9.2 \\
\#6 & BPO + ICL & BPO & \textbf{68.8} & 2.5 & 28.7 & \textbf{71.1} & 19.4 & 9.5 \\ \cmidrule{2-9}
\#7 & OD + ICL & OD & \textbf{60.0} & 2.5 & 37.5 & \textbf{52.4} & 9.1 & 37.3 \\ 
\#8 & BPO + ICL & BPO & \textbf{68.8} & 2.5 & 28.7 & \textbf{71.1} & 19.4 & 9.5 \\
\#9 & OD + ICL & BPO + ICL & \textbf{75.0} & 11.3 & 13.7 & \textbf{68.9} & 21.4 & 10.7 \\ \bottomrule
\end{tabular}
}\caption{Comparison of different Black-Box optimization methods across Vicuna Eval and Self-Instruct based on querying GPT-4.}
\label{ablation_study_icl}
\end{table}

\subsection{Ablation Study}
To further illustrate the advantages of our proposed self-instructed RL method (OD) and the importance of the ICL environment (ICL), we use derived prompts generated by Llama3-8B to query GPT-4, where the generated response are then evaluated by GPT-4. Tab.~\ref{ablation_study_icl} presents the comparison various combinations of OD and ICL with other methods using two evaluation datasets: Vicuna Eval~\cite{chiang2023vicuna} and Self-Instruct~\cite{wang2022self}.
By comparing \#1 and \#2, it can be observed that from the perspective of prompt refinement, our OD already exhibits higher quality than BPO, thereby promoting LLMs to generate more helpful responses. In Vicuna Eval, OD's win rate is $67.6\%$, significantly higher than BPO's $53.8\%$ and those are $50.8\%$ and $44.9\%$ in Self-Instruct Eval, where we achieve $5.9\%$ performance improvement. The comparison between OD and BPO in \#3 also support this analysis.

In \#5, when BPO is augmented by ICL, it significantly improves performance, achieving a win rate of $65.0\%$ in Vicuna Eval and $48.0\%$ in Self-Instruct. This is a noticeable improvement over BPO alone (\#4), which has a win rate of $53.8\%$ in Vicuna Eval and $44.9\%$ in Self-Instruct. Furthermore, when BPO + ICL is compared directly with BPO (\#6), BPO + ICL achieves a win rate of $66.3\%$ and $51.7\%$ in two datasets, demonstrating our proposed ICL query formulation is a general framework suitable with various prompt refinement methods.  

Finally, \#7, \#8, and \#9 examine the combination of OD with ICL. When OD is combined with ICL and BPO (\#7 and \#8), significant higher win rates demonstrate that our proposed ICL querying framework is suitable to various methods and promptes LLMs generate more helpful response. When compared with BPO + ICL (\#9), OD + ICL achieves an impressive win rate of $63.9\%$ in Vicuna Eval and $47.5\%$ in Self-Instruct, suggesting that the OD + ICL combination consistently provides stable and superior performance improvements across different environments.

In summary, our derived prompts enhanced by self-instructed RL are more effective than BPO prompts, in promoting LLMs to produce high-quality responses. Additionally, our proposed ICL querying formulation is a flexible, effective, and general framework that can enhance various prompt refinement methods in obtaining better responses to the original queries.

\begin{figure}[h]
    \centering
    \includegraphics[width=0.8\linewidth]{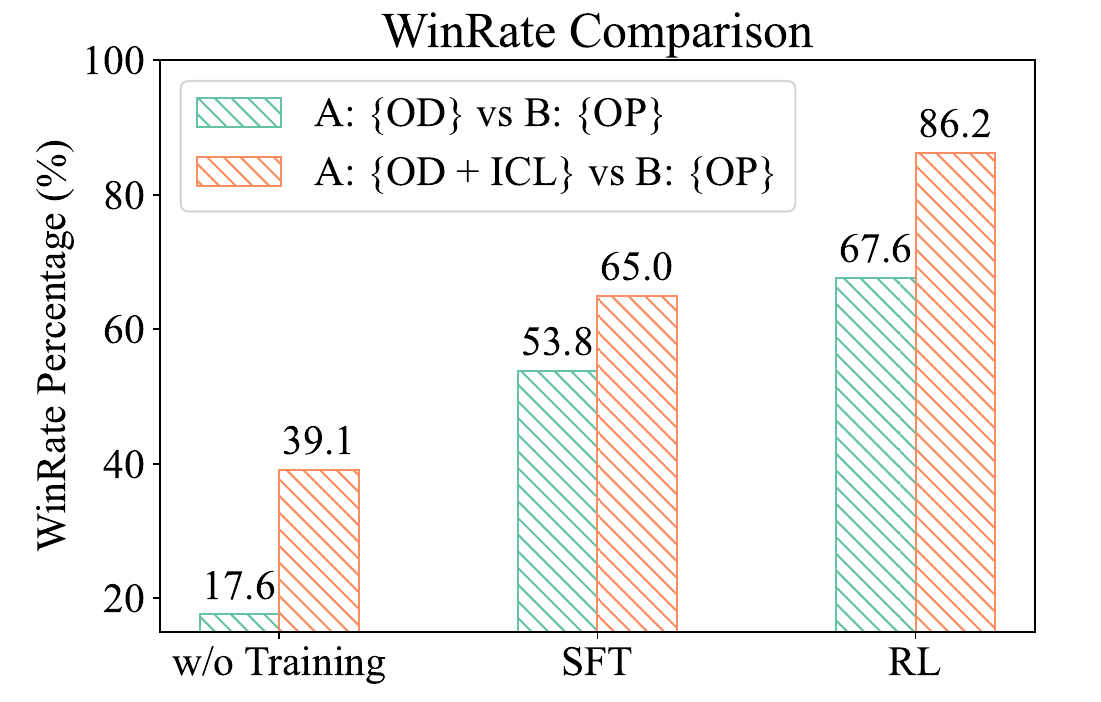}
    \caption{Win rates comparison in Vicuna Eval among w/o training, SFT and RL. Based on Llama3-8B for generation of derived prompt.}
    \label{fig:rl_ablation}
\end{figure}

\begin{figure*}[t]
    \centering
    \includegraphics[width=1\linewidth]{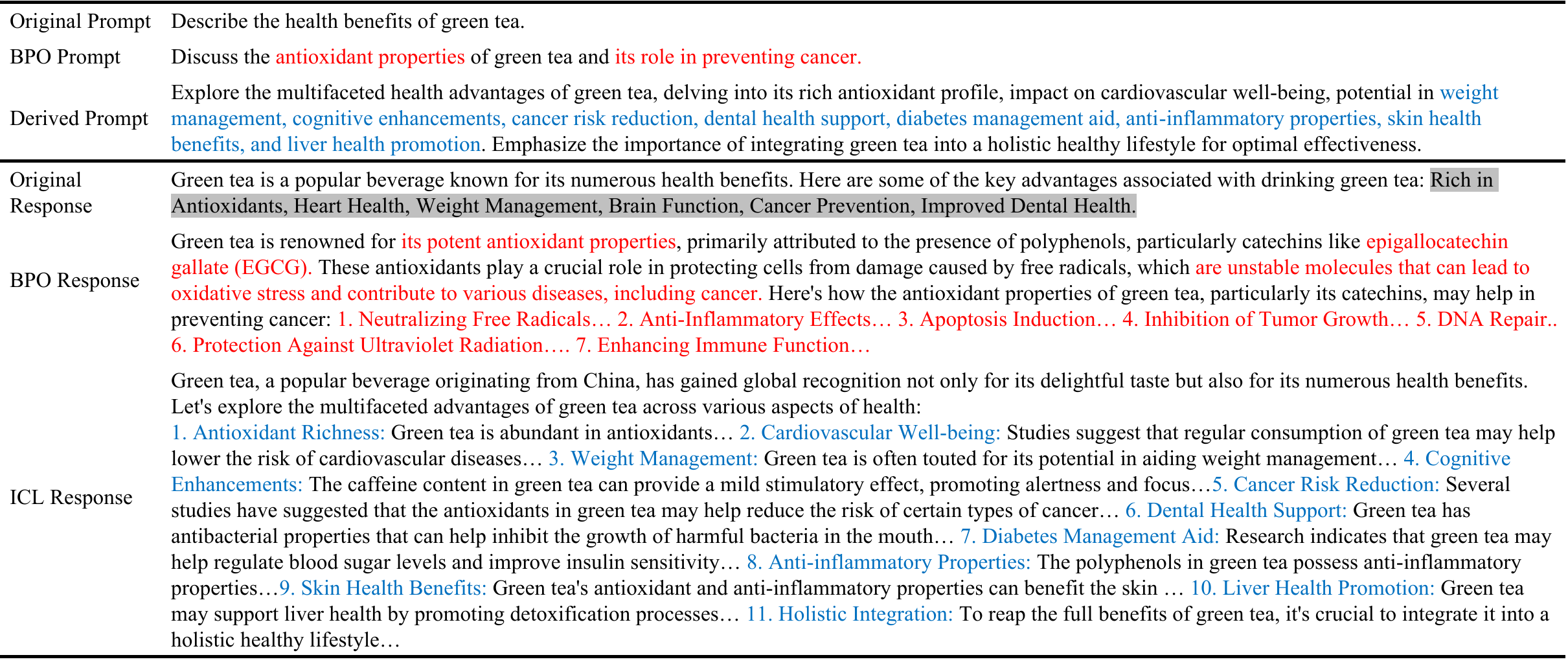}
    \caption{Detailed cases study. We compare quality of different types of responses by querying GPT-4.}
    \label{fig:case_study}
\end{figure*}

\subsection{Independent Analysis of Self-Instructed RL and ICL Query Framework}
In this section, we mainly focus on independently analyzing how our proposed self-instructed RL and ICL query framework perform. Fig.~\ref{fig:rl_ablation} shows two advantages of our methods:

\textbf{Effectiveness of ICL framework:} Without training, directly using LLM to generate derived prompts combined with ICL (OD + ICL) has already achieved excellent results, achieves $39.1\%$ win rates improvement. This supervising performance indicates that our proposed ICL framework is highly general and effective even without additional optimization. 

\textbf{Effectiveness of self-instructed RL:} After optimization with RL, our methods achieves a higher win rates increasing from $17.6\%$ to $67.6\%$ on {OD vs OP}, and that from $39.1\%$ to $86.2\%$ on {OD + ICL vs OP}, which is also better than using SFT. This highlights the significant advantage of our proposed RL in generating higher quality derived prompts.

\subsection{Case Study}
As shown in Fig~\ref{fig:case_study}, BPO significantly alters the original prompt's intent, restricting the response to the relationship between green tea and cancer while neglecting other benefits. In contrast, our derived prompt maintains consistency with the original prompt's content and expands upon it. Consequently, the comparison between the original response and the ICL response reveals that our approach not only effectively covers the information present in the original response, but also stimulates the LLM's intrinsic knowledge, resulting in more comprehensive and detailed descriptions.



\section{Related Work}
\paragraph{Prompt Refinement}
Prompts are crucial for guiding LLMs to produce more helpful response, leading to extensive research on improving their quality. Initially, prompt optimization relied on manually crafted templates~\cite{reynolds2021prompt}, a labor-intensive process with interpretation challenges~\cite{webson2021prompt}. Recent studies automate this process using techniques like gradient-based search~\cite{shin2020autoprompt, pryzant2023automatic}, paraphrasing~\cite{haviv2021bertese, jiang2020can} and leveraging LLMs to generate prompts~\cite{zhou2022large, fernando2023promptbreeder, yang2024zhongjing, cheng2023black}. Additionally, RL based methods are designed to optimize a prompt rewritten model through reward functions and task-specific templates~\cite{deng2022rlprompt,kong2024prewrite,zhang2022tempera,huang2024enhancing}.

\paragraph{RLHF}
RLHF has been widely explored to align LLMs with human preferences~\cite{stiennon2020learning,ouyang2022training,bai2022constitutional,lee2023rlaif}. Common approaches include building a reward model using maximum likelihood estimation (MLE) and optimizing it with the Proximal Policy Optimization (PPO) algorithm~\cite{schulman2017proximal}. However, replicating PPO's success has proven challenging for the open-source community due to the high resource demands. To address this, some research has shifted to offline direct preference learning~\cite{zhao2023slic,rafailov2024direct,li2023remax}, which bypasses reward modeling and directly optimizes a loss target using an offline dataset. Among these methods, ours and~\citet{cheng2023black} fall into better alignment with Black-Box model.

\paragraph{Improving In-Context Learning}
Several approaches have been introduced to enhance in-context learning (ICL) performance by improving the selection of in-context examples. Some methods focus on refining template selection~\cite{yin2023did}, while others aim to enhance the choice of examples~\cite{liu2021makes,rubin2021learning}. Additionally,~\citet{wan2023better} introduces a criterion for evaluating examples based on some criteria. Other recent innovations include flipped learning~\cite{ye2022guess} and noisy channel prompting~\cite{min2021noisy}. For multiple-choice problems,~\citet{xu2023k} suggests using K-nearest neighbors for label assignment, and~\citet{yang2023iterative} proposes iterative context updates. 


\section{Conclusion}

In this paper, we introduced an innovative method for enhancing LLM performance using an automatically generated in-context learning framework. By creating derived prompts through a self-instruct RL mechanism, our approach enriches the context of the original prompts. Extensive experiments reveal that our framework significantly improves response quality, even for Black-Box models like GPT-4. Excellent performance suggests that our method offers a promising paradigm for aligning LLMs without modifications, thereby improving their overall usability and effectiveness in various applications.


\bibliography{aaai25}

\clearpage
\appendix
\onecolumn
\begin{center}
        \LARGE\textbf{Supplementary Materials For \\ Self-Instructed Derived Prompt Generation Meets In-Context Learning: Unlocking New Potential of Black-Box LLMs}
\end{center}
\section{Failed Refined Prompt Cases}
In this section, we provide more failed refined prompt cases. In this situation, directly querying the response model (e.g., GPT-4) with the refined prompt would definitely obtain useless response 

\begin{table*}[h]
\centering
\resizebox{1\textwidth}{!}{
\begin{tabular}{l|l}
\toprule
Original Prompt& Compose a rap verse about a topic of your choosing.\\ 
Refined Prompt& Compose a rap verse about the topic of \rr{Artificial Intelligence (AI).} \\ \midrule

Original Prompt& \makecell[l]{Write a tweet that is opinionated and captures \bb{the essence of the following text: The use of artificial intelligence}\\ \bb{is fundamentally changing the way that business is done.}}\\ 
Refined Prompt& Write a tweet \rr{that expresses an opinion about how artificial intelligence is revolutionizing business.}\\ \midrule

Original Prompt&What was the average house price in San Francisco?\\ 
Refined Prompt&What was the average house price in San Francisco \rr{in 2020?}\\ \midrule

Original Prompt& Open up a discussion about a given topic.\\ 
Refined Prompt& \makecell[l]{Let's discuss the current and future impact of \rr{artificial intelligence (AI) on society. Consider its applications,}\\ \rr{potential benefits, drawbacks, and ethical considerations.}}\\ \midrule

Original Prompt&why do people from barcelona talk so weird?\\ 
Refined Prompt&What are some characteristics of the Barcelona accent?\\ \midrule

Original Prompt&what is the Adam\\ 
Refined Prompt&Who is Adam \rr{in the Bible?}\\ \midrule

Original Prompt&Tell me about Hong Kong\\ 
Refined Prompt&\makecell[l]{Tell me about Hong Kong's \rr{response to COVID-19. (Please provide specific information on measures taken} \\ \rr{by the government, such as border control, mask mandates, and vaccination efforts.)}} \\ \midrule

Original Prompt&Who is JFK?\\ 

Refined Prompt&Who is JFK?\rr{ Who is the current president of the United States?}  Provide accurate and detailed responses.\\
\bottomrule
\end{tabular}
}
\caption{Failed refined prompt cases.}
\label{more_failed_prompts}
\end{table*}

\section{Instruct-guided Derived Prompt Generation Template $x_{\text{DPG}}$}
Recall in Section Self-instructed RL for Derived Prompt Generation Model, we design a derived prompt generation instruction $x_{\text{DPG}}$ to ask the model $\pi_{\theta}$ to generate a derived prompt, instead of using a $\pi_{\text{SFT}}$. Here we proved the specific template shown as below:

\begin{tcolorbox}
\begin{flushleft}
\textit{\#\#\# Instruction: Please provide a more comprehensive, easily understandable, and answerable version of the following question. Ensure that necessary contextual information is added during the rewrite, but do not limit the understanding and response to the question. Avoid confining the question to just a few aspects, allowing the responder to think from multiple angles. Only return the refined question and do not explain. Here is my original question:". } \\
\hspace*{\fill}\\
\hspace*{\fill}\\
\textit{\#\#\# Question: \{Original Prompt $x$\}. } \\

\end{flushleft}
\end{tcolorbox}
As shown in Alg.~\ref{alg:alg_1}, we firstly concatenate $x_{\text{DPG}}$ with original prompt $x$ together as $X$, which will be then sent to $\pi_{\theta}$ to obtain the derived prompt $x'$.

\section{Experiments}
\paragraph{Training Settings.} For training settings, we use ReMax~\cite{li2023remax} to train the derived prompt generation model facilitated by DeepSpeed ZeRO-2~\cite{deepspeed2024}. We set the temperature parameter to $\tau = 1.0$ and use nucleus sampling with a parameter of $\text{top}_p = 0.9$ for all models. The maximum length for derived prompt generation and response model is set to 256 tokens. We conduct experiments on 4 NVIDIA A100 GPUs. All experiments are trained with a learning rate of $1 \times 10^{-6}$ for 2 epoch with decay, where the KL penalty $\beta$ is set to 0.05 for all models. Our batch size is set to 1.

\paragraph{Reward Evaluation} In addition to GPT-4 as evaluator for comparing the quality of response promoted by different methods, we also adopt Reward Score into quantitative evaluation, which provides more comprehensive understanding to our methods. Therefore, we train a Llama3-8B-Instruct as derived prompt generation model and then query GPT-4. As shown in Tab.~\ref{ablation_study_reward}, we can observe our method can prompote GPT-4 generate more useful and higher quality response in the view of Reward Score, except for Dolley Eval.

\begin{table*}[h]

\centering
\resizebox{0.5\textwidth}{!}{
\begin{tabular}{c|c|c|c}
\toprule
Dataset & Method & Reward Score & $\delta$ \\ \midrule
\multirow{3}{*}{Vicuna} & OURS & \textbf{3.97} & - \\
 & OP & 3.72 & \textcolor{ggg}{$\uparrow$ 0.25} \\
 & BPO & 3.83 & \textcolor{ggg}{$\uparrow$ 0.14} \\ \midrule
\multirow{3}{*}{BPO-Text Eval} & OURS & \textbf{3.19} & - \\ 
 & OP & 1.89 & \textcolor{ggg}{$\uparrow$ 1.30} \\
 & BPO & 2.42 & \textcolor{ggg}{$\uparrow$ 0.77} \\ \midrule
\multirow{3}{*}{Dolley Eval} & OURS & 3.24 & - \\ 
 & OP & 2.05 & \textcolor{ggg}{$\uparrow$ 1.19} \\
 & BPO & \textbf{3.25} & \textcolor{red}{$\downarrow$ -0.01} \\ \midrule
\multirow{3}{*}{Self-Instruct Eval} & OURS & \textbf{2.79} & - \\
 & OP & 0.72 & \textcolor{ggg}{$\uparrow$ 2.07} \\
 & BPO & 1.78 &  \textcolor{ggg}{$\uparrow$ 1.01} \\ \bottomrule
\end{tabular}
}\caption{Comparison of different Black-Box optimization methods across Vicuna Eval and Self-Instruct based on querying GPT-4.}
\label{ablation_study_reward}
\end{table*}

\section{GPT-4 Original Prompt-based Justification Prompt}
In this section, we give a GPT-4 based pair-wise justification prompt shown as below.

\begin{tcolorbox}
\begin{flushleft}
\textit{
\#\#\# System message: \\
Please act as an impartial judge and evaluate the quality of the responses provided by two AI assistants to the user question displayed below. You should choose the assistant that follows the user's instructions and answers the user's question better. Your evaluation should consider factors such as the helpfulness, relevance, accuracy, depth, creativity, and level of detail of their responses. Begin your evaluation by comparing the two responses and provide a short explanation. Avoid any position biases and ensure that the order in which the responses were presented does not influence your decision. Do not allow the length of the responses to influence your evaluation. Do not favor certain names of the assistants. Be as objective as possible. After providing your explanation, output your final verdict by strictly following this format: '[[A]]' if assistant A is better, '[[B]]' if assistant B is better, and '[[C]]' for a tie.}
\hspace*{\fill}\\
\textit{Prompt template:} \\
\textit{\{User Question\}} \\ 
\textit{\text{\{Original Prompt\}}}\\
\hspace*{\fill}\\
\textit{[The Start of Assistant A's Answer]}\\
\textit{\text{\{answer a\}}}\\
\textit{[The End of Assistant A's Answer]}\\
\hspace*{\fill}\\
\textit{[The Start of Assistant B's Answer]}\\
\text{\{answer b\}}\\
\textit{[The End of Assistant B's Answer]}\\

\end{flushleft}
\end{tcolorbox}

\section{More Case Study}
In this section, we provide two more examples shown as below.

\begin{figure}[t]
    \centering
    \includegraphics[width=\linewidth]{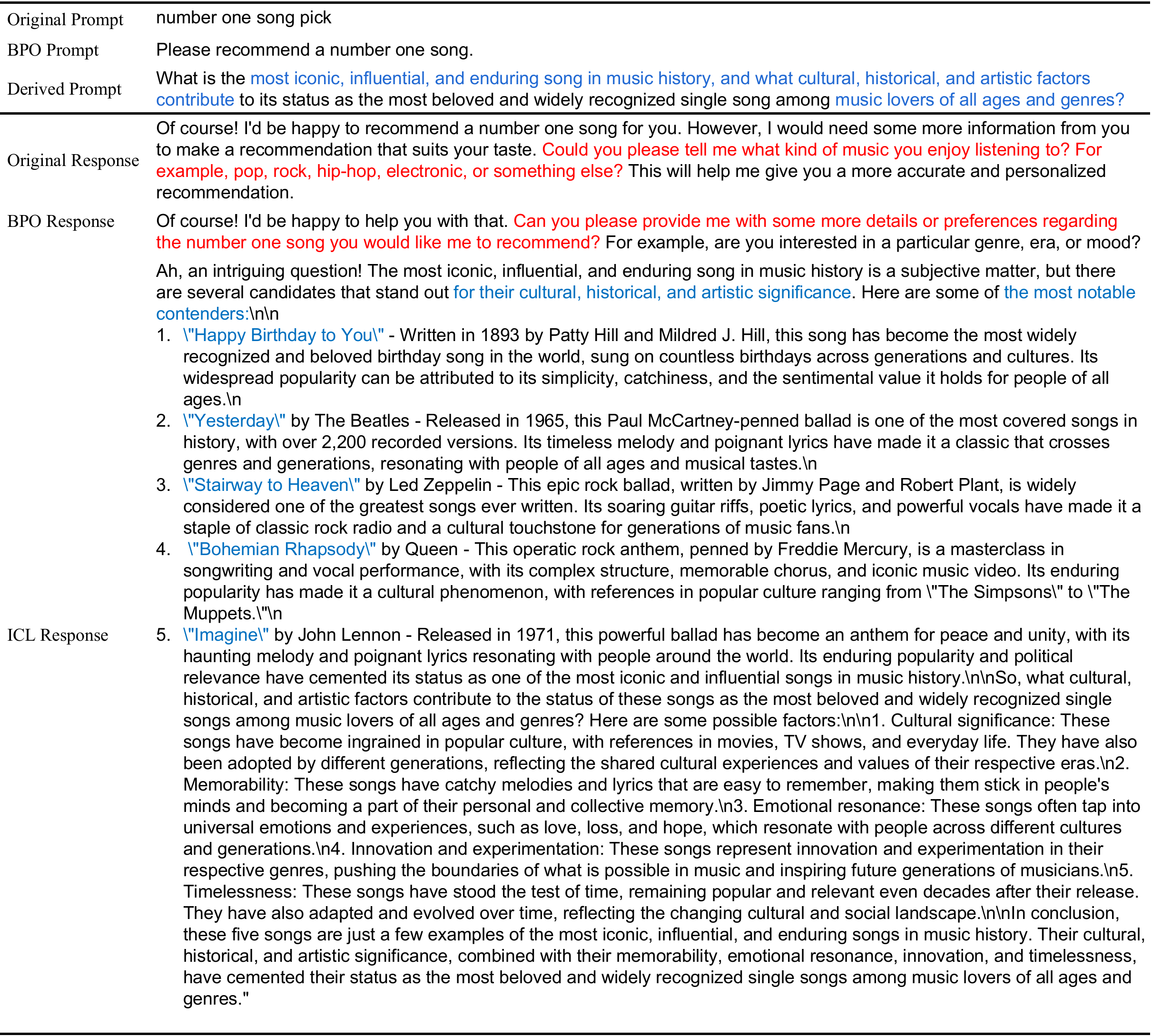}
    \caption{}
    \label{fig:more_cases_1}
\end{figure}

\begin{figure}[t]
    \centering
    \includegraphics[width=\linewidth]{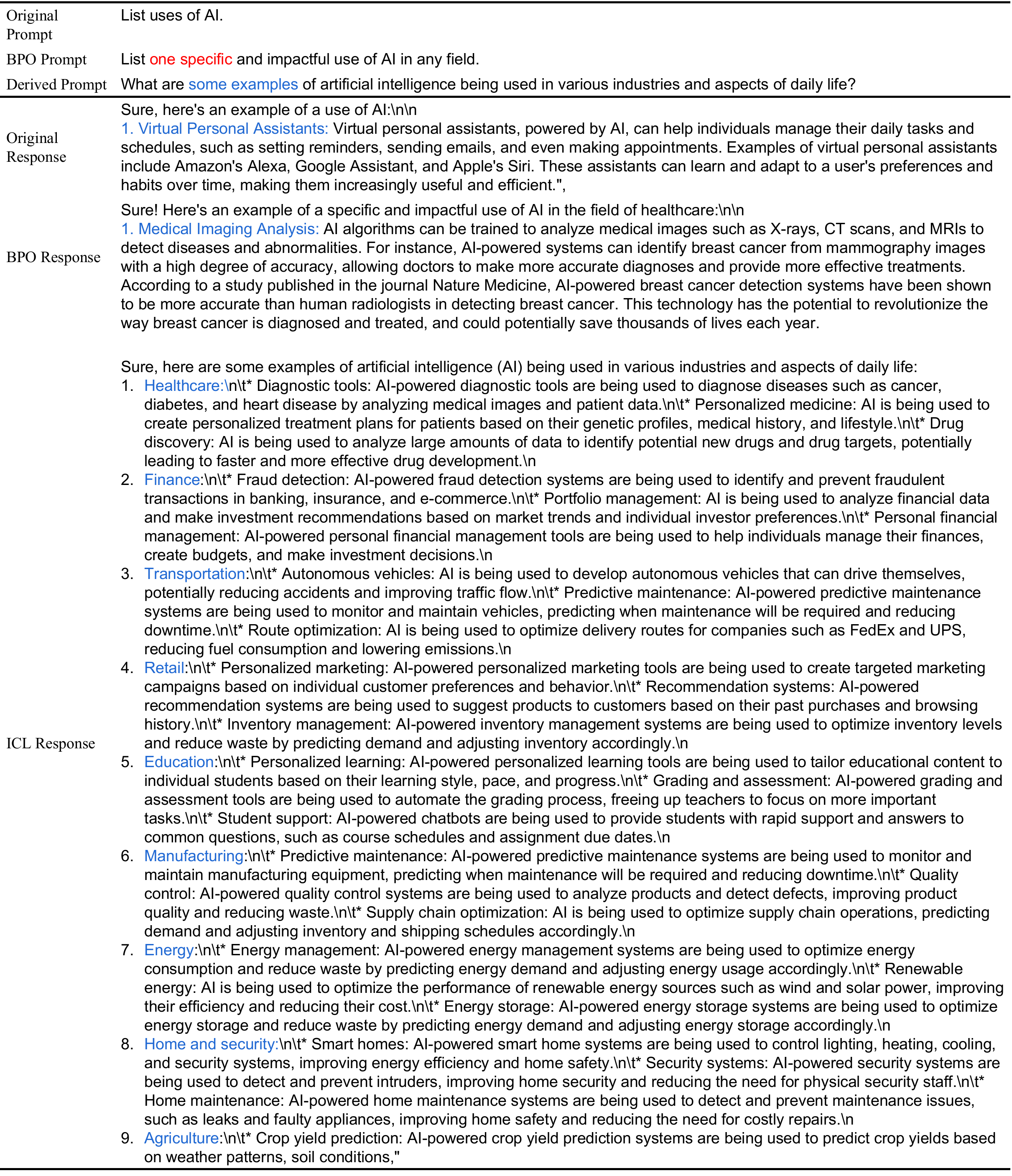}
    \caption{}
    \label{fig:more_cases_2}
\end{figure}
\end{document}